\documentclass[12pt,twoside]{article}
\usepackage{latexsym,amsmath,amssymb,hyperref,graphicx,algorithm2e}
\def\eqvsp{}  \newdimen\paravsp  \paravsp=1.3ex
\newtheorem{theorem}{Theorem}
\newtheorem{lemma}[theorem]{Lemma}
\newenvironment{keywords}{\centerline{\bf\small
Keywords}\begin{quote}\small}{\par\end{quote}\vskip 1ex}

\def\eqvsp{}  \newdimen\paravsp \paravsp=1.3ex

\def\,{\mskip 3mu} \def\>{\mskip 4mu plus 2mu minus 4mu} \def\;{\mskip 5mu plus 5mu} \def\!{\mskip-3mu}
\def\dispmuskip{\thinmuskip= 3mu plus 0mu minus 2mu \medmuskip=  4mu plus 2mu minus 2mu \thickmuskip=5mu plus 5mu minus 2mu}
\def\textmuskip{\thinmuskip= 0mu                    \medmuskip=  1mu plus 1mu minus 1mu \thickmuskip=2mu plus 3mu minus 1mu}
\def\dispmuskip{}\def\textmuskip{}
\textmuskip
\def\beq{\eqvsp\dispmuskip\begin{equation}}    \def\eeq{\eqvsp\end{equation}\textmuskip}
\def\beqn{\eqvsp\dispmuskip\begin{displaymath}}\def\eeqn{\eqvsp\end{displaymath}\textmuskip}
\def\bqa{\eqvsp\dispmuskip\begin{eqnarray}}    \def\eqa{\eqvsp\end{eqnarray}\textmuskip}
\def\bqan{\eqvsp\dispmuskip\begin{eqnarray*}}  \def\eqan{\eqvsp\end{eqnarray*}\textmuskip}

\def\paradot#1{\vspace{\paravsp plus 0.5\paravsp minus 0.5\paravsp}\noindent{\bf\boldmath{#1.}}}

\def\req#1{(\ref{#1})}

\def\nq{\hspace{-1em}}
\def\qed{\hspace*{\fill}\rule{1.4ex}{1.4ex}$\quad$\\}
\def\fr#1#2{{\textstyle{#1\over#2}}}
\def\SetR{I\!\!R}
\def\SetN{I\!\!N}
\def\SetZ{Z\!\!\!Z}
\def\qmbox#1{{\quad\mbox{#1}\quad}}
\def\v{\boldsymbol}
\def\trp{{\!\top\!}}
\def\s{\sigma}
\def\t{\theta}
\def\tr{\mbox{tr}}

\begin{document}

\title{\vspace{-4ex}
\vskip 2mm\bf\Large\hrule height5pt \vskip 4mm
Matching 2-D Ellipses to 3-D Circles with \\
Application to Vehicle Pose Identification
\vskip 4mm \hrule height2pt}
\author{{\bf Marcus Hutter} and {\bf Nathan Brewer} \\[3mm]
\normalsize RSISE$\,$@$\,$ANU and SML$\,$@$\,$NICTA \\
\normalsize Canberra, ACT, 0200, Australia \\
\normalsize \texttt{nbrewer@cecs.anu.edu.au \ \ marcus@hutter1.net}
}
\date{December 2009}
\maketitle

\begin{abstract}
Finding the three-dimensional representation of all or a part of a
scene from a single two dimensional image is a challenging task. In
this paper we propose a method for identifying the pose and location
of objects with circular protrusions in three dimensions from a
single image and a 3d representation or model of the object of
interest. To do this, we present a method for identifying ellipses
and their properties quickly and reliably with a novel technique
that exploits intensity differences between objects and a geometric
technique for matching an ellipse in 2d to a circle in 3d.

We apply these techniques to the specific problem of determining the
pose and location of vehicles, particularly cars, from a single
image. We have achieved excellent pose recovery performance on
artificially generated car images and show promising results on real
vehicle images. We also make use of the ellipse detection method to
identify car wheels from images, with a very high successful match
rate.
\def\contentsname{\centering\normalsize Contents}
{\parskip=-2.7ex\tableofcontents}
\end{abstract}


\newpage
\section{Introduction}

Determining three-dimensional information from a single
two-dimensional image is a common goal in computer vision \cite{Hartley2004}.
Circular features appear frequently in real world objects, the mouth
of a cup, the wheels of a car, the iris of a human eye. Such
circular features (referred to as 3d circles from here
on) appear as ellipses when imaged if viewed non-frontally.

In this paper we present a method for matching parts of a
two-dimensional scene to existing three-dimensional models using a
technique that maps three-dimensional circles to two-dimensional
ellipses. Additionally, we present a novel method for identifying
elliptical regions and the properties thereof in a two-dimensional
image. In particular, we aim to recover the pose of motor vehicles
in natural scenes. The purpose of this is twofold: First, it allows
us to determine the location and facing of a car in a street scene,
which has obvious uses in driver assistance systems, and second it
allows us to place vehicles in three dimensions from a single image
without constructing a depth map of the entire scene. It is
important to note that this paper does not present a general-purpose
vehicle detection or complete scene reconstruction algorithm, rather
it aims to deliver a basis from which such platforms can be built or
improved.

Intelligent driver assist technology is becoming more and more
important as our roads become busier. Determining vehicle pose in
three dimensions allows us to estimate a heading for a visible
vehicle, which can alert the driver to approaching vehicles.
Importantly, our pose detection technique requires only a
single frame, allowing it to detect stationary vehicles as potential
hazards in case of unexpected motion, unlike many techniques which
require a series of images of moving vehicles.

Developing a depth map of a scene from a single image is an
underconstrained problem. There are approaches based on machine
learning \cite{Saxena07}, or that make use of the effects of fog
\cite{Tan08}, but these either require classification and a large
training data set or specific scene conditions. Our method is able
to place 3D models of imaged vehicles at their locations in the same
orientation as in the real world. This has applications
for security, as observers will be able to determine the position of
a vehicle from surveillance footage, as well as scene
reconstruction. Unlike the work of Hinz et al. \cite{Hinz03}, which
places simple 3d car models in vehicle locations in aerial
photographs, we want to use a specific, detailed 3d model to match
to street-level car images.

In order to do this, we first must isolate ellipses in the image
that we can then match to circular areas in a three-dimensional
model. For vehicles, the most applicable choice for these ellipses
are the wheels. Existing ellipse detection methods, such as the
Hough transform \cite{Tsuji78} require a five-dimensional search,
which has high computational and memory requirements. There are many
recent methods methods which aim to reduce this by using
parametrised spaces for the Hough transform accumulator, such as
\cite{Chia07, McLaughlin1998}. These methods still require
edge extraction or other image preprocessing, and often detect
significantly more ellipse than we need in this paper. To this
end, we present a new method for identifying wheels in real-world
vehicle images which incorporates a significant amount of ellipse
filtering and parameter estimation as part of the detection
process. Other feature detection methods are discussed more fully in
the text.

While this work is focussed on identifying the pose of cars, the
ellipse to circle mapping process can be applied to any three
dimensional object with circular features, as shown in the
experimental verification of Section \ref{2d3dmatch}.

\section{2D Ellipse Detection}

\begin{figure*}[t!]
\begin{center}
\includegraphics[width=\textwidth]{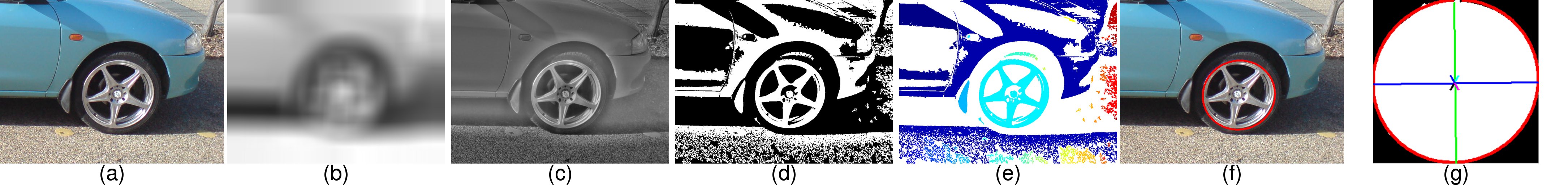}
\caption{Images showing each step of the wheel detection Algorithm
1. (g) shows the identified wheel ellipse with the
major and minor axes shown in blue and green respectively. the four
lines in the center represent the potential $\v\varphi$ normal
vectors, scaled 50 times for clarity. Note that each $\v\varphi$
vector has a large component pointing into the page.}
\label{fig:ellipsealgorithm}
\end{center}
\end{figure*}

Rather than using an existing ellipse detection method based on edge
detection, such as those presented in \cite{Mai07, Kenichi02}, we
propose a method for detecting elliptical objects which allows for
significant noise at the edge of the object. The ellipse detection
method presented here involves several steps. First, pixels which
are brighter than their surroundings are identified by comparing
each pixel to the average local intensity. Connected components, or
blobs, are then found and labeled. Further processing is performed
on each of these blobs to fill gaps and then to determine whether
the detected object is elliptical. While we use bright image regions
here, the method can be easily changed to identify dark regions or
regions which adhere to some other color or intensity property. This
process is summarised in Algorithm 1 and visually in
Figure \ref{fig:ellipsealgorithm}.

\begin{algorithm}[t!]
\caption{\small The Ellipse Detection Algorithm. For more details, see text
and Figure \ref{fig:ellipsealgorithm}.} \label{alg:detect}
\begin{small}
\SetLine
\KwData{A greyscale or full-colour image (Figure \ref{fig:ellipsealgorithm}a)}
\KwResult{Location and parameters of ellipses in the image (Figure \ref{fig:ellipsealgorithm}f)}
\nl Smooth image with square kernel (Figure \ref{fig:ellipsealgorithm}b)\;
\nl Normalize Image (Figure \ref{fig:ellipsealgorithm}c)\;
\nl Threshold Normalised Image (Figure \ref{fig:ellipsealgorithm}d)\;
\nl Detect and label connected regions (Figure \ref{fig:ellipsealgorithm}e)\;
\nl Star Fill connected regions (Figure \ref{fig:ellipsealgorithm}g)\;
\nl Get Ellipse parameters using Eq.(\ref{ellipseparams})\;
\nl Filter out non-ellipses by comparison to expected ellipse (Figure \ref{fig:ellipsealgorithm}g)\;
\nl Determine Ellipse Normal using Eq.(\ref{CtoPhi}) (Figure \ref{fig:ellipsealgorithm}g)\; \vspace{1ex}
\end{small}
\end{algorithm}

\paradot{Identifying Comparatively Bright Pixels}
In order to identify pixels which are brighter than their
surrounding area, a new image which reflects the local mean
intensity for each pixel is built. Each pixel in this new image
contains the mean intensity for a square region centered on that
pixel in the original image. This square is set to \textit{b}\% of
the shortest image side. For the purposes of this paper, a window
10\% of the width of the image was found to be effective, but this
can depend on the nature of the ellipses that are desired.
If this window would extend beyond the edges of the image for a
particular pixel, the mean of the closest possible window which is
entirely contained within the image is used instead.
The intensity integral image \cite{ViolaJones} is used to
efficiently calculate this windowed mean value, which was the primary
motivation for using a simple box kernel for smoothing.

We then subtract this average from the intensity of each pixel in
the initial image, and then set a threshold \textit{T} which defines
how much a pixel must exceed the local average to be flagged as
`bright'. For wheel detection, one standard deviation above the mean
of the average image was found to produce good results. This
thresholding produces a binary image, with zero corresponding to
`dark' pixels, and one corresponding to `bright' pixels.

\begin{figure}[t!]
\begin{center}
\includegraphics[width = \textwidth]{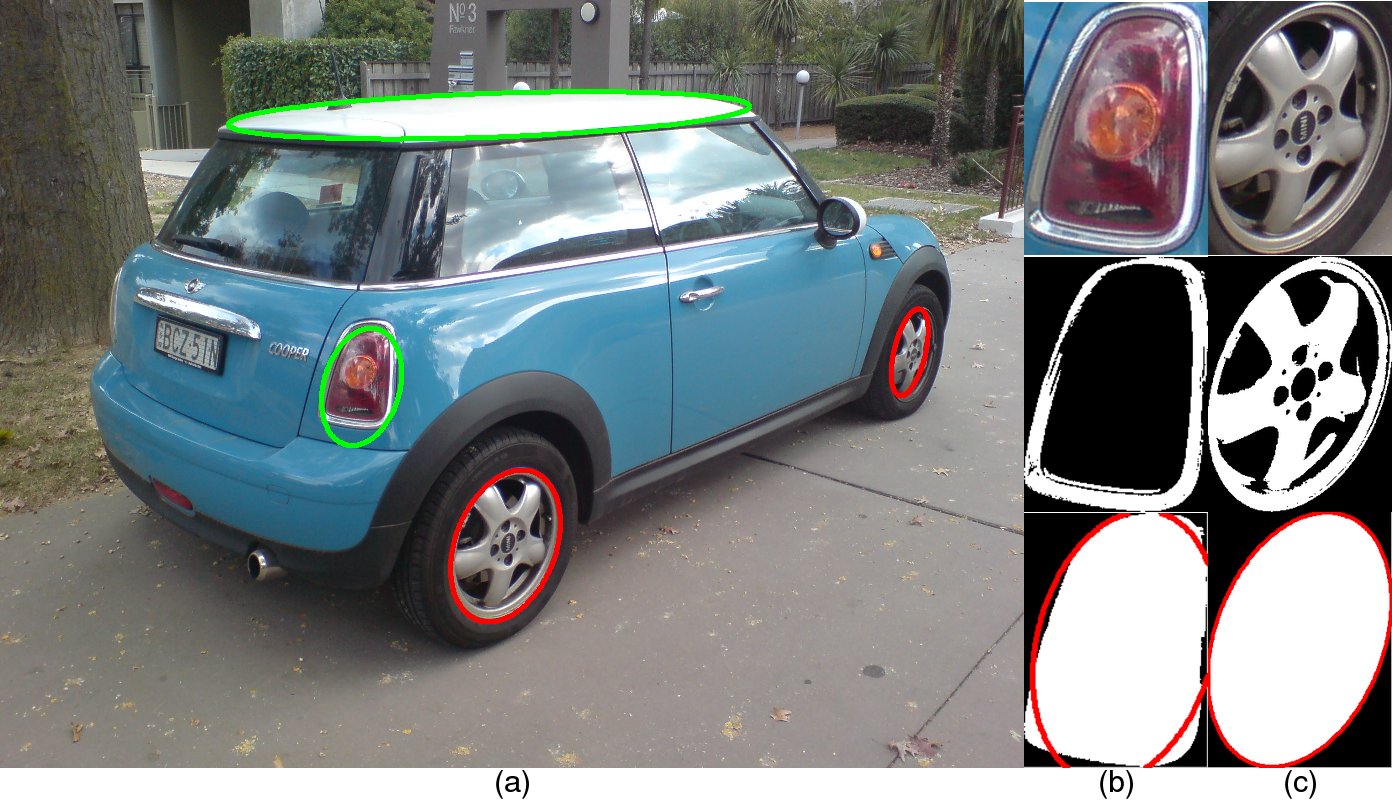}
\caption{(a) An image of a Mini Cooper showing in blue all detected
ellipses and in red the ellipses detected as wheels (b and c) zoomed
sections of the car containing a wheel and a non-wheel region,
respectively. (i) Zoomed region of image, (ii) the pixels identified
as `bright', (iii) Filled blobs and corresponding expected ellipses
in red} \label{fig:ellipsedetect}
\end{center}
\end{figure}

This thresholding method tends to produce a thick `edge' around the
ellipse perimeter as seen in Figure \ref{fig:ellipsedetect} (ii),
and frequently contains small gaps corresponding to darker areas of
the ellipse surface, corresponding to occlusions, decoration or
simply image noise.

\paradot{Labeling and Filling Connected Regions}
To denoise the binary image obtained above, it is first necessary to
eliminate small bright regions. Simple blob extraction is then
performed, labeling each 8-connected region with a unique
identifier.

In this ellipse detection method, objects being investigated must be
filled. As it is not desirable to affect the overall shape of a
blob, the fill method chosen preserves the perimeter and fills gaps
in a radial fashion. In each blob, if a pixel is marked as bright
then all pixels on a line between it and the blob center are also
marked as bright. We are able to do this efficiently, in linear
time, by spiralling in from the edge to the centre. This fill method
is used rather than a simple flood fill as blobs frequently do not
have fully enclosed edges due to noise or occlusions. Each of these filled, denoised blobs is then
separately investigated to determine if it is elliptical in nature.
See the following section and Figure \ref{fig:ellipsedetect} (iii)
for more details.

\paradot{Extraction of Blob Properties and Ellipse Comparison}
In order to determine whether a particular blob is elliptical, the
extracted blob is compared to an equivalent ellipse. We do this by
finding the mean and covariance matrix of each blob. Let $W
\subset\SetZ\times\SetZ$ be the pixels of one. The area
$|W|\in\SetN$, mean $\v\mu\in\SetR^2$ and covariance matrix
$C\in\SetR^{2\times 2}$ of each blob can then be found as follows:
\bqan\label{ellipseparams}
  |W| = \sum_{\v p\in W} 1,\qquad
  \mu \;=\; {1\over|W|}\sum_{\v p\in W}\v p,\\
  C = {1\over|W|}\sum_{\v p\in W}(\v p-\v\mu)(\v p-\v\mu)^\trp
\eqan
By assuming that the blob is an ellipse, the major and minor axes of
the corresponding ellipse can be calculated from the eigenvalues
$a_1$ and $a_2$ of the blob covariance matrix, and the center of the
corresponding ellipse is given by $\v\mu$. From this information, we
are able to find the shape of the corresponding ellipse by
Eq.\req{ellipsefromparams}. This approach is similar to that described in \cite{Safaee91}. This step is shown in Figure \ref{fig:ellipsedetect} (iii). We are able to perform a consistency
check at this stage to eliminate some obvious non-ellipses: If $W$
is an ellipse, its area is given by $\pi\mathrm{ab}$ as well as
$|W|$. Thus if $||W|-4\pi\sqrt{\mathrm{det}C}|>T$ for some small
threshold $T$, we can immediately discard this blob.

We can then directly compare the remaining blobs to the equivalent
ellipse on a per pixel basis by counting the number of pixels in
which the blob and its ellipse do not match. As we are interested in
the elliptical nature of the blob regardless of size, we divide the
number of mismatches by the blob area. Blobs with a sufficiently low
mismatch can be regarded as ellipses. We found experimentally that a
mismatch ratio of 20\% identifies human-visible ellipses while
discarding non-ellipses.

\section{Mapping a 2D Ellipse to a 3D circle}\label{2d3dmatch}

Circles in 3d space become ellipses in 2d space after parallel
projection. It is possible to exploit this fact to determine the
pose of a 3d object containing circular features from a projected
image.

\paradot{Describing an Ellipse by its Covariance Matrix}
First, we describe an ellipse in a way that is easily manipulated. Let
\beq\label{ellipsefromparams}
  E(\v\mu,C) \;:=\; \{\v p\in\SetR^2 : (\v p-\v\mu)^\trp C^{-1}(\v p-\v\mu)\leq 4\}
\eeq
describe a solid ellipse with center $\v\mu$, where $C$ is a symmetric positive definite matrix. It can be shown by integration that $C$ is the covariance of the ellipse given by $E(\v\mu,C)$. The eigenvalues $\lambda_1>\lambda_2$ of $C$ can be computed from the trace $\tr C=\lambda_1+\lambda_2$ and the determinant $\det C=\lambda_1\lambda_2$:
\beq\label{evalue}
  \lambda_{1/2} = \fr12\tr C \pm \sqrt{(\smash{\fr12}\tr C)^2-\det C}
\eeq
Therefore, the lengths of the major and minor axes of the ellipse are
\beq\label{axes}
  a_1=2\sqrt{\lambda_1} \qmbox{and} a_2=2\sqrt{\lambda_2}
\eeq

\paradot{Transforming a 3D circle to a 2D ellipse}
In Lemma \ref{circletoellipse}, we describe how a 3d circle projects
to an ellipse in 2d.

\begin{lemma}\label{circletoellipse}
Let
\beqn
  \text{Circle}:=\{\v x\in\SetR^3: ||\v x||\leq r \;\;\&\;\; \v x^\trp\v\varphi=0\}
\eeqn
be a centered circle with radius $r$ and unit normal $\v\varphi\in\SetR^3$.
Let $E(\v\mu,\v C)$ be the orthogonal projection of Circle along the $z$ axis. Then
$r=a_1$ and $\v\varphi$ and $C$ are related by
\beq\label{CtoPhi}
  \v\varphi \;\equiv\;
  \left(\begin{array}{c}\varphi_x\\ \varphi_y\\ \varphi_z\end{array}\right)
  \;=\; {1\over a_1}\left(\begin{array}{c}\pm\sqrt{a_1^2-4C_{xx}}\\ \pm\sqrt{a_1^2-4C_{yy}}\\ \pm a_2\end{array}\right)
\eeq
where $a_1$ and $a_2$ are defined in \req{axes} and \req{evalue}.
\end{lemma}

\paradot{Proof} The $(x,y)$ coordinates of $\v x=(x,y,z)\in$Circle describe an
ellipse. Condition $\v x^\trp\v\varphi=0$ implies
$z\varphi_z=-x\varphi_x-y\varphi_y$. Inserting this into $||\v x||$
yields
\bqa
  ||\v x||^2 &=& x^2+y^2+z^2
 \; = \;x^2+y^2+\left({x\varphi_x+y\varphi_y\over\varphi_z}\right)^2\notag\\
  &=& \binom{x}{y}^\trp\binom%
    {1+\varphi_x^2/\varphi_z^2\quad \varphi_x\varphi_y/\varphi_z^2}%
    {\varphi_x\varphi_y/\varphi_z^2\quad 1+\varphi_y^2/\varphi_z^2}
    \binom{x}{y}
\eqa
Hence the projected circle is an ellipse with covariance
\bqa\label{phitoC}
  C \;\equiv\; \binom{C_{xx}\;C_{xy}}{C_{yx}\;C_{yy}}
  &=& {r^2\over 4}\binom{1+\varphi_x^2/\varphi_z^2\quad \varphi_x\varphi_y/\varphi_z^2}%
    {\varphi_x\varphi_y/\varphi_z^2\quad 1+\varphi_y^2/\varphi_z^2}^{-1}\notag\\
    &=& {r^2\over 4}\binom{1-\varphi_x^2\quad -\varphi_x\varphi_y}{-\varphi_x\varphi_y\quad 1-\varphi_y^2}
\eqa

where we have exploited $||\v\varphi||=1$ when inverting the matrix.
Conversely, given an ellipse with covariance $C$, using
\req{evalue}, \req{axes}, and \req{phitoC}, the radius and normal of
the underlying 3d circle are given by \req{CtoPhi}.\qed

There is one hard sign constraint, namely
$\mbox{sign}(\varphi_x\varphi_y)=-\mbox{sign}C_{xy}$. Since a
projected circle is usually only visible from one side of the 3d
object, we define $\v\varphi$ as pointing away from the observer into
the image. This fixes the sign of $\pm a_2$, leaving two solutions
which can not be disambiguated. These solutions arise from the fact
that an ellipse can be projected from a circle facing in two
different directions.

\paradot{Accuracy}
We give some rough estimates of the accuracy of the reconstructed
$\v\varphi$. In particular we show that reconstruction from a frontal
view is somewhat surprisingly harder than from a tilted view.

Let $\t=\arccos|\varphi_z|\in[0;\pi/2]$ be the angle by which the
Circle is tilted out of the $(x,y)$-plane, and let $\Delta a_2$ be
the accuracy to which the minor axis $a_2$ could be determined by
the ellipse detection Algorithm 1. For ``perfect'' pixel ellipses, an
accuracy between $\Delta a_2\approx 1$ (pixel accuracy) and
$\Delta a_2\approx 1/a_2$ (sub-pixel accuracy due to
averaging/anti-aliasing) is achievable. From
$a_2=r|\varphi_z|=r\cos\t$, we get by Taylor expansion,
\beqn
  \Delta a_2 \;=\; -r\sin\t\,(\Delta\t) - \fr12 r\cos\t(\Delta\t)^2 + O((\Delta\t^3))
\eeqn
Hence for a non-frontal view ($\t\gg\Delta\t$), the orientation of
the circle axis $\v\varphi$ can be determined with accuracy $\Delta\t\approx
|\Delta a_2|/r\sin\t$, while for a frontal view ($\t\ll\Delta\t$), the
accuracy reduces to $\Delta\t\approx\sqrt{2|\Delta a_2|/r}$.
This is consistent with our experimental results on real cars like
those in Figure \ref{fig:fullpageresults}.

\paradot{Determining the Mapping from a Circle to an Ellipse}
We now want to determine the orthogonal projection that maps a
general circle in 3d to a general ellipse in 2d.
Let $\v\nu,\v\phi\in\SetR^3$, and $R>0$ be the center, normal, and
radius of the circle, let $\v\mu\in\SetR^2$ and $C\in\SetR^{2\times
2}$ be the center and covariance matrix of the ellipse, and let
\beq\label{project}
  \v x' = \sigma Q\v x+\v q
\eeq
be the orthogonal projection to be determined. Matrix
$Q\in\SetR^{2\times 3}$ consists of the first two rows of an
orthogonal matrix $\tilde Q\in\SetR^{3\times 3}$, $\sigma>0$ is a
scaling factor, and $\v q\in\SetR^2$ a shift vector. The centers
$\v\nu$ and $\v\mu$ must match, hence $\v q=\v\mu-\sigma Q\v\nu$.
The major ellipse axis $a_1$ corresponds to the circle radius $R$ by
\req{CtoPhi}, hence $\sigma=a_1/R$. Let $\v\varphi$ be the
``normal'' of the ellipse as defined in Lemma \ref{circletoellipse}.
Note that there are two possibilities for this vector, as there is a
sign ambiguity in \req{CtoPhi}. We use the convention that
$\v\varphi$ and $\v\phi$ point from ellipse center away from the
observer ($\varphi_z<0$). Axle $\v\phi$ ``projects/rotates'' to
$\tilde Q\v\phi$, which must align with $\v\varphi$, hence we have
the constraint $\v\varphi=\tilde Q\v\phi$ on $\tilde Q$ (and hence
$Q$).

An orthogonal matrix has three degrees of freedom (dof), the
constraint $\v\varphi=\tilde Q\v\phi$ determines 2 dof, hence 1 dof
remains, namely after rotating $\v\phi$ into $\v\varphi$, we can
freely rotate around $\v\varphi$.

A rotation around a unit axle $\v u\in\SetR^3$ by angle $\alpha$ can
be represented by quaternion \cite{Forsyth03}
\bqa\label{quaternion}
  q \;:=\; a + bi + cj + dk \;\equiv\; a+(b,c,d)\notag\\
  \;:=\; \cos\fr\alpha2 + \v u\sin\fr\alpha2, \quad||\v u||=1
\eqa
which has the following matrix representation
\beq\label{qtomat}
 \left(
   \begin{array}{ccc}
     a^2\!+\!b^2\!-\!c^2\!-\!d^2 & 2bc\!-\!2ad & 2ac\!+\!2bd \\
     2ad\!+\!2bc & a^2\!-\!b^2\!+\!c^2\!-\!d^2 & 2cd\!-\!2ab \\
     2bd\!-\!2ac & 2ab\!+\!2cd & a^2\!-\!b^2\!-\!c^2\!+\!d^2 \\
   \end{array}
 \right)
\eeq
In our case, a rotation around $\v\phi\times\v\varphi$ by angle
$\alpha=\cos^{-1}(\v\phi\circ\v\varphi)\in[0;\pi]$ rotates $\v\phi$
into $\v\varphi$. Using
\begin{align}\label{sincoshalf}
  &\cos\fr\alpha2=\sqrt{\smash{\fr12}(1+\cos\alpha)} \qmbox{and}\notag\\
  &\sin\fr\alpha2=\mbox{sign}(\alpha)\sqrt{\smash{\fr12}(1-\cos\alpha)}
  \qmbox{for} \alpha\in[-\pi;\pi]
\end{align}
leads to the following convenient representation:
\bqa\label{phitovarphi}
  a = \cos\fr\alpha2 = \sqrt{\smash{\fr12}(1+\v\phi\!\circ\!\v\varphi)},
  (b,c,d) = \v u\sin\fr\alpha2 = \fr1{2a}(\v\phi\!\times\!\v\varphi)
\eqa
This representation only breaks down if $\v\phi$ is exactly
antiparallel to $\v\varphi$. In this case, any $\v u$ orthogonal to
$\v\phi$ will do.

Subsequently, a rotation around $\v u=\v\varphi$ by an arbitrary
angle $\beta$ can be applied. Hence the complete transformation is as follows:

\begin{lemma}\label{onewheelalign}
Eq.\req{project} projects 3d circle
$(\v\nu,\v\phi,R)$ = (center,axle,radius) to 2d ellipse
$(\v\mu,C)$ = (center,covariance) iff
\bqan
  Q = \mbox{first 2 rows of } \tilde Q,
  \tilde Q = \tilde Q_2\tilde Q_1,
  \v q = \v\mu-\sigma Q\v\nu,
  \sigma = a_1/R,
\eqan
Minor axis $a_1$ is given by \req{evalue} and \req{axes}.
Matrix $\tilde Q_1$ is given by \req{qtomat} and \req{phitovarphi} and \req{CtoPhi}.
Matrix $\tilde Q_2$ is given by \req{quaternion} and \req{qtomat}
with arbitrary $\alpha=\beta\in[-\pi;\pi]$ but now with
$\v u:=\v\varphi$ given by \req{CtoPhi}.
\end{lemma}

\paradot{Experimental Verification}\label{2d3dmatch:eg}
%
\begin{figure}[t!]
\begin{center}
\includegraphics[width=0.7\textwidth]{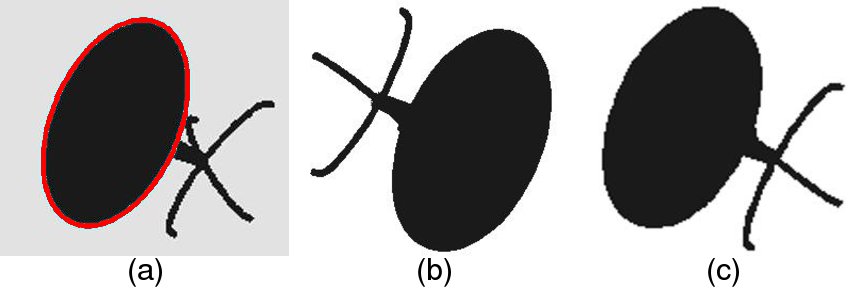}
\caption{An image which shows (a) an image rendered from a 3d model of a table with the ellipse to be matched highlighted in red and (b,c) the possible model poses determined by the algorithm}
\label{fig:CircleToEllipse}
\end{center}
\end{figure}
\begin{table}
\caption{The true and estimated pose of a simple 3d
table.}\label{tab:simpletest}\vspace{-1ex}
\begin{center}
\begin{small}
\begin{tabular}{ccc|ccc}
\multicolumn{3}{c}{True Pose}&\multicolumn{3}{|c}{Estimated Pose}\\
Axle $\v\varphi$ & $\nq$Scale $\s\nq$ & Shift $\v q$ & Axle $\v\varphi$ & $\nq$Scale $\s\nq$ & Shift $\v q$ \\
\hline\hline
[0.7568 0.3243 -0.5676] & 6 & [400 340] & [0.7296 0.3193 -0.6048] & 6.06 & [405 340]\\
\end{tabular}
\end{small}
\end{center}
\end{table}
In order to test this 2d-3d matching procedure, we generated test
images from a simple 3d table model to determine its accuracy. The
table was arbitrarily rotated, scaled and shifted (see Figure
\ref{fig:CircleToEllipse}) and then projected to produce a 2d
image.
The center ($\v\mu$), major and minor axis lengths ($a_1$ and $a_2$
respectively) and covariance ($C$) of the projected ellipse
corresponding to the top of the table were determined using the
ellipse detection method presented above. This information and
Eq.\req{CtoPhi} are used to estimate the normal to this
ellipse. Lemma \ref{onewheelalign} was then used to estimate $Q$,
$\sigma$ and $\v q$, which could then be compared to the input
rotation, shift and scale. Table \ref{tab:simpletest} shows the
recovered and input values for the table, and Figure
\ref{fig:CircleToEllipse} displays graphically the initial and
recovered poses. Note that the only the two sets of results produced by the
sign ambiguity in \req{CtoPhi} are presented in this figure. We have assumed that the normal points towards the page.

\section{2D Wheel Detection}

In order to match a 3d model of a vehicle to a 2d image with an
ellipse to circle mapping technique, it is first necessary to find
the ellipses corresponding to circular features on the vehicle in
the image. Wheels are used for this purpose, as they are both easily
locatable in the 2d representation and are present in all vehicles.
This also allows for the matching of an image to a generic vehicle,
should a specific model be unavailable.

For the purposes of this section, it is assumed that the image being
investigated features a vehicle as its primary object. It is also
assumed that the image contains the two wheels on the same side of
the vehicle clearly, and that the vehicle is upright in the image.
This image can be a segment of a larger image which an existing
vehicle detection system, such as \cite{Moutarde07}, has identified
as a vehicle. We make use of wheels because they have a number of exploitable properties. If the wheels are not visible in the input image, an alternative technique must be used.

A typical car has hubcaps that are significantly brighter than the
surrounding tyre, appearing as a comparatively bright, unobstructed
ellipse. This allows us to use the ellipse detection method
presented above, with a few additional steps to filter out detected
non-wheel ellipses. While it is the case that some vehicles have unusually dark hubcaps or rims, simple observation of vehicles in the street shows that this is far from the norm, allowing our method to applicable in the majority of cases where a side view of a vehicle is available.

This procedure identifies all `bright' ellipses in the image, and
can identify road markings, headlights and other circular regions in
the image. As we want only to identify the wheels, we need to
identify the two ellipse that are most likely to be wheels. To do
this, we make a few additional assumptions about the way a vehicle
appears in an image. First, in a vehicle the wheels cannot be too
large or small, so filled blobs containing less than 0.15\% or more
than 25\% of the image area are excluded.

The remaining blobs are then examined in a pairwise fashion to
determine the relative ellipse orientation of each pair and the
orientation of a line drawn between the centers of each ellipse in
the pair. Pairs are excluded as wheels if the ellipses have a
significantly different orientation or if the line connecting their
centers deviates more from the horizontal than is expected in a
natural image. If this process results in more than a single set of
wheels being detected, the pair with the lowest mismatch ratio are
identified as wheels. The properties of these ellipses can then be
matched to the circular wheels of a 3d car model, as described in
Sections 3 and 5.

There are other methods which can be used to identify the wheels in
a vehicle image. For example, a cascaded classifier system similar
to that used for face detection \cite{ViolaJones} could be adapted
to find wheels in images, just as it has been adapted to detect road
signs \cite{niklas_iv2008_1}. Similarly, a robust feature
description for a wheel could be developed using SIFT \cite{Lowe99}
or sift like features, which could be used to find wheel locations.
These approaches suffer from two major drawbacks which our approach
does not. First, these methods require labelled training data to
learn model parameters, which requires significant amounts of
human input. Additionally, the hubcaps/rims of almost every vehicle
are at least subtly different, and the direction from which they are
imaged has a large impact on their appearance, and so constructing a
training set with sufficient detail becomes even more arduous.
Second, these approaches only identify the location of a wheel, and
further processing must be done to extract the ellipse parameters
that we require for 2d-3d matching.

\section{Determining Vehicle Pose from a Single Image}

In the previous section, we have presented a method for both
identifying the wheels in a 2d image of a vehicle and determining
their ellipse properties. In this section, we present a method for
using these properties together with those from section
\ref{2d3dmatch} to determine the 3d pose of a vehicle from a single
2d image. It is possible to determine the location and orientation
of a vehicle to within a plane of rotation using the method
presented in section \ref{2d3dmatch}. We perform this matching on
the rear wheel, as the front wheel might be rotated out of plane
with the rest of the vehicle, which results in poor pose matching.
As input images contain two ellipses, corresponding to each of the
two wheels of the vehicle, it is possible to use information about
the second ellipse to eliminate this rotational ambiguity.
Additionally, information about the second wheel can improve the
scale factor $\sigma$ of the transformation.

We determine the ``normal'' of the ellipse for both wheel ellipses
in the 2d image using the method shown in Section \ref{2d3dmatch}.
The parameters $\sigma$ and $\beta$ in Lemma \ref{onewheelalign} are
found by matching the center $\v\nu'$ of the 3d model front wheel to
the corresponding ellipse center $\v\mu'$ in the image.

The 2d ellipse detection cannot determine which is the front and
which the back wheel, and also does not know which side of the car
the two wheels belong to. Together with the two solutions from
\req{CtoPhi}, we hence have a discrete 8-fold ambiguity. We are able
to eliminate four of these by assuming that the vehicle in the image
is upright, which means that the rear wheel on the left hand side of
the model matches the left hand wheel in the image if the car is
oriented with the front towards the right of the image, and the 3d
right rear wheel matches the right-hand image ellipse.

First, we determine a more accurate scale factor by finding the
length $||\v\Delta||$ of the vector $\v\Delta$ between the front and rear wheels in
the 3d model. This distance is invariant to model
rotation, and so we are able to find this easily. We then estimate
the 3d distance between wheels in the 2d image by setting
$(\delta_x,\delta_y)^\trp:=\v\mu'-\v\mu$ as the vector from the back
to the front wheel in the image. We know that $\v\delta\in\SetR^3$
should be orthogonal to $\v\varphi$, which gives us
$\delta_z=-(\delta_x\varphi_x+\delta_y\varphi_y)/\varphi_z$. We then
set $\sigma=||\v\delta||/||\v\Delta||$.

\begin{lemma}\label{twowheelalign}
The 3d back wheel with center $\v\nu\in\SetR^3$ and axle
$\v\phi\in\SetR^3$ and 3d front wheel with center $\v{\nu'}$ project
to 2d ellipses with center $\v\mu\in\SetR^2$ and covariance matrix
$C\in\SetR^{2\times 2}$ and center $\v{\mu'}\in\SetR^2$ respectively by
Lemma \ref{onewheelalign} with the following adaptation:
Scale $\s:=||\v\delta||/||\v\Delta||$ and $\beta$ is now constrained by
\beq\label{twowheelrotation}
  \cos\beta={\v\delta\circ\v\Delta\over ||\v\delta|| ||\v\Delta||},\quad
  \sin\beta={\det(\v\delta,\v\Delta,\v\varphi)\over ||\v\delta|| ||\v\Delta||}
\eeq
\bqan
 &(\delta_x,\delta_y)^\trp := \v\mu'-\v\mu, \quad
   \delta_z := -(\delta_x\varphi_x+\delta_y\varphi_y)/\varphi_z,\\
 & \v \Delta = \tilde Q_1(\v\nu'-\v\nu).
\eqan
\end{lemma}

\paradot{Proof} First we align the back wheel using Lemma \ref{onewheelalign}.
We then find the vector from the back to the front
wheel of the image in both the 3d model and the 2d image. Let
$\v\Delta:=\tilde Q_1(\v\nu'-\v\nu)$ be the vector from back to
front wheel in 3d after rear wheel alignment. We assume that
$\v\Delta$ is orthogonal to the rear wheel axle $\tilde
Q_2\v\phi\equiv\v\varphi$. The 2d rear wheel to front wheel vector
is given by $\v\delta$ above.

$\v\Delta$ must scale and rotate around $\v\varphi$ into $\v\delta$,
which is achieved by scaling with $\sigma$ and rotating with $\tilde
Q_2$ as defined by Lemma \ref{onewheelalign} and
$\beta$ given by \req{twowheelrotation}. \qed
This process leaves us with a 4-fold ambiguity which can be resolved
using a loss-based disambiguation. This disambiguation will be
discussed in a future paper.
We are also able to make use of
numerous consistency checks at this stage. As we do not use the size
of either wheel, we are able to compare the major axis length of the
imaged wheels with the 3d wheel radii. Also the front and back wheel
axles $\v\varphi$ and $\v{\varphi'}$ and distance $\v\delta$ must
(approximately) lie in a plane.

\section{Experimental Results}

The 2d ellipse to 3d circle matching method presented here was
tested on a number of images, both real and artificial. We have used
three 3d car models, sourced from linefour \cite{linefour} in our
experiments, a Volkswagon Golf, a BMW M3 and an Audi A6. Real images
were taken of vehicles in a number of locations around the
university under natural conditions. Where possible, we have matched
real images to the most similar 3d model available.

We manually select the pose with the best match from
the four generated by our method when producing output images. In addition, we have
developed a sophisticated loss-based method for automatically
selecting the best pose, but this is outside the scope of this
paper.

We found that the parallel projection model that we have assumed works well even for perspective projection, provided that the perspective effect is not too extreme. This is due to the fact that each wheel is considered separately when determining ellipse parameters, and each occupies only a small volume of 3d space, resulting in minor length distortion which still produces good performance.

\paradot{Testing Dataset}
Artificial images were generated by rendering 3d models using freely
available MATLAB toolkits. Renders were generated with a plain
background at a resolution of 800x600. Real images were collected
using a variety of hand held cameras. Images were rescaled where necessary to provide a consistent image resolution.

\begin{figure}[t!]
\begin{center}
\includegraphics[width=\textwidth]{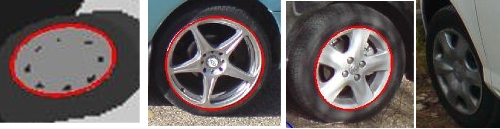}
\caption{Close ups of wheels from an artificial image (a) and real images (b,c) with the equivalent ellipse in red. (d) an example of a wheel for which detection fails.}
\label{fig:wheelmatchzooms}
\end{center}
\end{figure}

\paradot{Wheel Detection Performance}
Detecting the ellipses in the image which correspond to the
vehicle's wheels and accurately identifying their properties is an
important step in vehicle pose recovery. The wheel detection
algorithm presented was tested on a number of images, both real and
artificial taken in a wide variety of locations ranging from
artificially lit underground car parks to outdoor on street parking.
15 artificial and 35 real-world images were sampled, containing
numerous different car models, lighting conditions and viewing
angles. Each image contained two unobstructed wheels that were
clearly identifiable by a human observer. Figure
\ref{fig:wheelmatchzooms} shows segments of the images which
contained wheels and the ellipses extracted by the wheel detection
algorithm.

The wheel detection algorithm works quite well in most cases,
however it has difficulty identifying wheel hubs which are quite
oblique to the viewer or is too dark to be clearly differentiated
from the tyre by the chosen threshold. Of the 50 vehicle images
tested, both wheels were identified correctly in 76\% of images,
while one wheel was identified in 20\%. In the two images for which
no wheels were detected, the wheel's hubcaps were unusually dark, as
in Figure \ref{fig:wheelmatchzooms}(d). Other wheel detection
failures occurred when there were occlusions, either from other
objects or parts of the vehicle being in the way or dark, dirty or
spoke-like rims/hubcaps. Several single wheel detections occurred
when the algorithm identified a wheel in a second vehicle present in
the image rather than detecting both wheels of a single vehicle.
The wheel detection algorithm also has difficulty identifying wheels
when the hubcap of the vehicle is obstructed by an object which is
also comparatively bright, as these objects will disrupt the shape
of the blob containing the wheel. Overall, the detection performance
is quite good, with at least one wheel being identified in 96\% of
test images.

\begin{figure}[t!]
\begin{center}
\includegraphics[width = \textwidth]{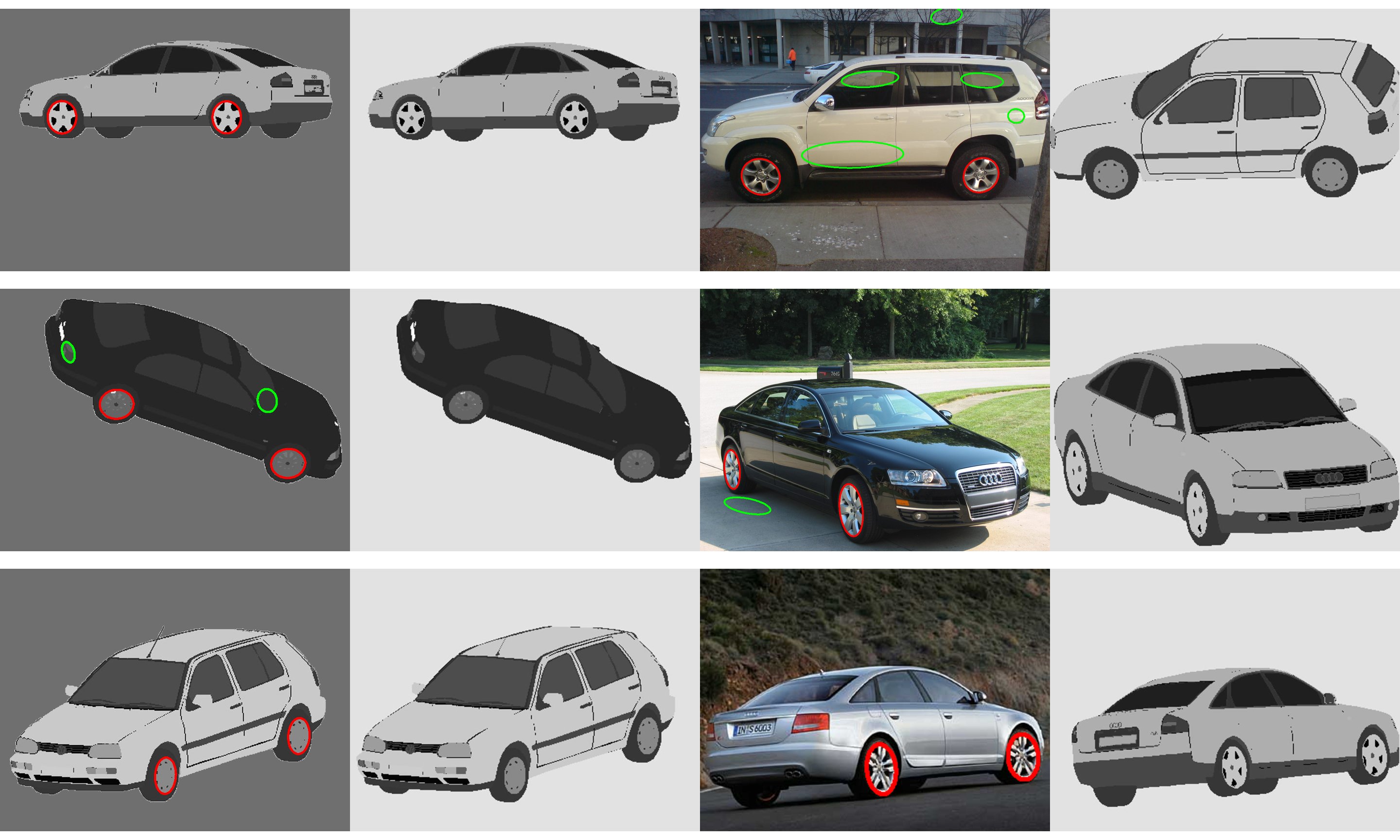}
\caption{A set of real and artificial images showing input images
with detected ellipses on the left and a model rotated to the
estimated 3d pose of the vehicle in the above image on the right
(4-fold ambiguity manually selected). Images cropped for
display purposes only. Note that in the original
photos used for ellipse detection, cars filled only a fraction
between 10\% and 25\% of the image. We cropped to the car region for
display purposes only.
} \label{fig:fullpageresults}
\end{center}
\end{figure}
To test the effect of scale on the wheel detection, a subset of
images were rescaled to 25, 50, 150, 200 and 400\% of the original
size. The algorithm detected the same ellipses in each of these
rescaled images, demonstrating that performance of wheel detection
is not greatly affected by scale.

\paradot{Artificial Images}
The pose recovery performance of the presented algorithm on
artificial images can be determined exactly, as we know the
manipulations performed on the model to produce the test image. We
perform two types of pose estimation on artificial images, one using
the wheel detection method presented in this paper and one using
exact ellipse normals determined exactly from the model and its
rotation parameters. This allows us to verify both the accuracy of
the pose estimation and the wheel detection steps.

\begin{table*}[t!]
\caption{Table of results showing input pose parameters and
estimated pose from ellipse normals derived from the wheel detection
algorithm. Pose from manually extracted normals matched the true
pose exactly, and so has not been
included}\label{tab:artificialresults}
\begin{center}
\begin{scriptsize}
\begin{tabular}{c|c c c|c c c}
Vehicle & \multicolumn{3}{c|}{True Pose}&\multicolumn{3}{|c}{Pose from Wheel Detection}\\
Model & Quaternion Rotation $q$ & $\nq$Scale $\s\nq$ & Shift $\v q$ & Quaternion Rotation $q$ & $\nq$Scale $\s\nq$ & Shift $\v q$\\
\hline\hline
Golf & -0.2162 -0.2162 -0.6053 0.7350 & 60 & [400 300] &  -0.2139 -0.2083 -0.6077 0.7359  & 61.03 & [396 297]\\
& -0.4417 -0.5522 0.5522 -0.4417 & 50 & [380 330] & -0.4400 -0.5315 0.5536 -0.4663 & 49.99 & [428 323]\\
\hline
Audi &0.9701 0.000 0.2425 0.000 & 150 & [400 300] & 0.9691 0.0070 0.2367 -0.0036 & 150.99 & [396 304]\\
& 0.9701 0.000 -0.2425 0.000 & 150 & [400 300] &  0.9690 0.0087 -0.2469 0.0007 & 149.78 & [389 304]\\
\hline
BMW & 0.0948 0.1896 -0.9481 0.2370 & 100 & [400 300] &  0.0852 0.1883 -0.9468 0.2466 & 99.34 & [402 299]\\
&  0.8729  -0.2182 0.4364  0.000  & 100 & [400 300] & 0.8755  -0.2212 0.4297 -0.0019 & 97.88 & [363 311]\\
\end{tabular}
\end{scriptsize}
\end{center}
\vspace{-3ex}
\end{table*}

Table \ref{tab:artificialresults} shows the true and estimated pose
using the wheel normal estimation method. The rotation is given as
quaternion parameters $a, b, c, d$, the shift as a 2-dimensional
vector $[x,y]$ representing the shift from car centre to image
origin, and the scale factor is a scalar. Figure
\ref{fig:fullpageresults} shows graphically the recovered pose for
two artificial images using approximated wheel normals. The pose
recovered from exact ellipse normals matched exactly the input pose, and so is not included.

The pose returned when ellipse normals are precisely known matches
perfectly with the expected pose of the image. However, the
estimated ellipse normals produce a pose which does not exactly
match that expected. The pose error is relatively small, and it
clearly results from inaccuracies in the estimation of ellipse
parameters from the wheel detection. Even when using estimated
parameters, the pose of the vehicle is still close enough to the
true pose that it provides a good starting point for techniques
which make use of other information about the vehicle to fine-tune
the estimation.

\paradot{Real Images}
Figure \ref{fig:fullpageresults} shows three real-world car images and
the estimated pose of a 3d model of these vehicles. As we do not
know the true vehicle pose in a real image, we are unable to conduct
a quantitative analysis of the recovered pose, and must instead rely
on visual inspection.

Car pose is recovered quite well in terms of scale and position, but
there is some misalignment in orientation.  This misalignment is
caused by a distortion of the detected wheel ellipses, resulting in
an imprecise normal recovery and thus a mismatched pose. The
recovered model pose is still able to give us significant
information about the position of the vehicle.

\section{Summary and Conclusion}

We have presented a method for mapping a two-dimensional ellipse to
a three-dimensional circle by estimating the normal of the ellipse
in three dimensions. Our results show that an ellipse can be matched
to a circle in three dimensions to a pose with eight discrete and
one continuous ambiguity. We also show that these ambiguities can be
reduced by using other information about the object to be matched or
the scene it is drawn from. In particular, vehicle pose can be
resolved to a four-fold discrete ambiguity by making use of both of
a vehicle's wheels and the fact that a vehicle is not transparent.
Forcing the vehicle to be upright reduces this to a twofold
ambiguity, given that we know which wheel in the image is the rear.
We are currently developing a dissimilarity measure that will be able to
automatically select the correct pose from these ambiguities by performing 
a comparison between each returned pose and the input image, but detail on 
this process is beyond the scope of this paper. It is based 
on a distance between images which is invariant under
lighting conditions and relatively insensitive to the image
background.

Additionally, we described a technique for wheel detection and wheel
parameter estimation using a local thresholding technique and blob
covariance. Our results show that this technique performs well
on a variety of real and synthetic vehicle images.


\begin{small}

\end{small}

\end{document}